\pdfoutput=1
\documentclass[11pt]{article}
\usepackage[final]{ACL2023}
\usepackage{times}
\usepackage{latexsym}
\usepackage[T1]{fontenc}
\usepackage[utf8]{inputenc}
\usepackage{microtype}
\usepackage{hyperref}
\usepackage{inconsolata}
\usepackage{graphicx}
\title{1024m at SMM4H 2024: Tasks 3, 5 \& 6 - Self Reported Health Text Classification through Ensembles}

\author{Ram Mohan Rao Kadiyala \\
  University of Maryland, College Park \\
  \texttt{rkadiyal@terpmail.umd.edu} \\\And
  M.V.P. Chandra Sekhara Rao \\
  RVR\&JC College of Engineering \\
  \texttt{mvpcs@rvrjc.ac.in} \\}

\begin{document}
\maketitle
\begin{abstract}
Social media is a great source of data for users reporting information regarding their health and how various things have had an effect on them. This paper presents various approaches using Transformers and Large Language Models and their ensembles, their performance along with advantages and drawbacks for various tasks of SMM4H'24 - Classifying texts on impact of nature and outdoor spaces on the author's mental health (Task 3), Binary classification of tweets reporting their children's health disorders like Asthma, Autism, ADHD and Speech disorder (task 5), Binary classification of users self-reporting their age (task 6).
\end{abstract}

\section{Introduction}
Social media has become a key way for people to share their experiences and feelings. This has opened up new opportunities for researchers to understand how different aspects of life affect our well-being. The paper explores three tasks of SMM4H 2024\citep{smm4h-2024-overview} - 4-way classification of texts based on effect of nature, outdoor spaces and activities on author's mental health (Task 3), Binary classification of texts reporting health disorders in author's child including ADHD, Autism, Asthma and Speech disorder (Task 5)\citep{info:doi/10.2196/50652}, Binary classification of texts self-reporting author's exact age directly / indirectly (Task 6).The paper explores usage of transformer models like RoBERTa\citep{liu2019roberta}, DeBERTa\citep{he2021deberta}, Longformer\citep{beltagy2020longformer} and LLs including both proprietary and open-source like GPT-4\citep{openai2024gpt4}, Claude-Opus\citep{opus2024}, Llama-3 8B\citep{touvron2023llama}, Mistral 7B\citep{jiang2023mistral}, Gemma 7B\citep{gemmateam2024gemma}, and ensembles along with advantages and drawbacks of each approach using the models. Similar previous works can be found in \citep{weissenbacher-etal-2022-overview}, \citep{magge-etal-2021-overview} and \citep{klein-etal-2020-overview}.

\section{Datasets}
The dataset for Task 3 consists of 3000 reddit posts from r/socialanxiety belonging to four classes based on self reported impact of outdoor spaces and activities on the author's mental health - 0: unrelated to the task, 1: had a positive impact, 2: is neutral or had no effect, 3: had a negative effect. The dataset for Task 5 consists of 9734 tweets belonging to two classes - 1: users reporting having a child having ADHD, Asthma, Autism or Speech disorder and the rest as class 0. Similarly for Task 6, the dataset of 21200 texts consists of both tweets and reddit posts from r/AskDocs for two classes - Class 1 being texts through which the author's current age in years may be determined and rest as Class 0. The distribution of labels for the three tasks can be seen in \autoref{table:1}, \autoref{table:2} and \autoref{table:3}.

\begin{table}[h]
\centering
\begin{tabular}{|c|c|c|c|}
\hline
         & \textbf{Training} & \textbf{Development} & \textbf{Testing} \\ 
\hline
Class 0  & 1131  & 377  & ?    \\
Class 1  & 160   & 54   & ?    \\
Class 2  & 395   & 131  & ?    \\
Class 3  & 114   & 38   & ?    \\ 
\hline
Total    & 1800  & 600  & 600  \\ 
\hline
\end{tabular}
\caption{Dataset split and class distribution : Task 3}
\label{table:1}
\end{table}

\begin{table}[h]
\centering
\begin{tabular}{|c|c|c|c|}
\hline
         & \textbf{Training} & \textbf{Development} & \textbf{Testing} \\ 
\hline
Class 0  & 5118  & 254  & ?     \\
Class 1  & 2280  & 135  & ?     \\ 
\hline
Total    & 7398  & 389  & 1947  \\ 
\hline
\end{tabular}
\caption{Dataset split and class distribution : Task 5}
\label{table:2}
\end{table}

\begin{table}[h]
\centering
\begin{tabular}{|c|c|c|c|}
\hline
         & \textbf{Training} & \textbf{Development} & \textbf{Testing}  \\
\hline
Class 0  & 5966  & 2435   & ?     \\
Class 1  & 2834  & 1765   & ?     \\
\hline
Total    & 8800  & 4200   & 8200  \\ 
\hline
\end{tabular}
\caption{Dataset split and class distribution : Task 6}
\label{table:3}
\end{table}

\section{Systems Description}
For Task 3, two approaches were tested. One where classification was done directly in a 4-way and the other where classification was done is two stages, this involved first classifying the text whether it is related to the task or not i.e class 0 or not and then classifying the effect on the user in the second stage. For Task 5 and 6 it was done directly as a binary classification task\footnote{Code available at: \href{https://github.com/1024-m/ACl-2024-SMM4H-Task-3-5-6}{https://github.com/1024-m/SMM4H-ACL-2024}}\footnote{Models available at: \href{https://huggingface.co/1024m}{https://huggingface.co/1024m}}. In LLM approaches, The proprietary versions were used as zero-shot and the rest of the LLMs were tested in a zero-shot and fine-tuned manner. Additionally they were tested in a two stage classification for Task 3. In the case of ensembles, It was through majority voting in a set of models, through and-rule for high precision requirement and through or-rule for high recall requirements. For Task 5 and 6, while using LLMs, classification was done by dividing the criteria into parts and aggregating the individual results. i.e In the case of Task 5, individual prompts test for each condition that needed to be satisfied to classify as positive and AND-rule is used for generating final label. Similarly OR-rule was used for Task 6. The performance of different approaches can be seen in \autoref{table:4}, \autoref{table:5} and \autoref{table:6}.The data during training was shuffled after every epoch and also internally in each mini-batch.

\begin{table}[t]
    \centering
    \begin{tabular}{|c|ccc|}
    \hline
                              & \textbf{F1} & \textbf{P} & \textbf{R} \\
    \hline    
        Bart-Large* (2-stage) & 0.673 & 0.666 & \textbf{0.687} \\
        Bart-Large (direct)   & 0.654 & 0.676 & 0.643 \\
        Bart-Large (2-stage)  & \textbf{0.679} & \textbf{0.677} & 0.682 \\
    \hline
        Mean                  & 0.519 & 0.565 & 0.538 \\
        Median                & 0.580 & 0.630 & 0.589 \\
    \hline
    \end{tabular}
    \caption{Precision, Recall and F1 on Test set compared to other participants : Task 3}
    \smallskip
    \small * indicates model is trained without using Dev set
    \label{table:7}
\end{table}

\begin{table}[t]
    \centering
    \begin{tabular}{|c|ccc|}
    \hline
                             & \textbf{F1} & \textbf{P} & \textbf{R} \\
    \hline
        Bart-Large* (direct) & 0.912 & 0.896 & \textbf{0.929} \\
        Bart-Large  (direct) & \textbf{0.918} & \textbf{0.923} & 0.912 \\
    \hline
        Mean                 & 0.822 & 0.818 & 0.838 \\
        Median               & 0.901 & 0.885 & 0.917 \\
    \hline
    \end{tabular}
    \caption{Precision, Recall and F1 on Test set compared to other participants : Task 5}
    \smallskip
    \small * indicates model is trained without using Dev set
    \label{table:8}
\end{table}

\begin{table}[t]
    \centering
    \begin{tabular}{|c|ccc|}
    \hline
                             & \textbf{F1} & \textbf{P} & \textbf{R} \\
    \hline
        Bart-Large (direct)  & \textbf{0.959} & \textbf{0.953} & \textbf{0.965} \\
        GPT-4 (and-rule)    & 0.922 & 0.895 & 0.951 \\
    \hline
        Mean                 & 0.924 & 0.924 & 0.926 \\
        Median               & 0.936 & 0.934 & 0.949 \\
    \hline
    \end{tabular}
    \caption{Precision, Recall and F1 on Test set compared to other participants : Task 6}
    \label{table:9}
\end{table}


\begin{table*}[!ht]
    \centering
    \begin{tabular}{|l|ccc|ccc|}
    \hline
          & \multicolumn{3}{|c|}{\textbf{Direct Classification}} & \multicolumn{3}{|c|}{\textbf{2-Stage Classification}} \\
    \hline
        \textbf{Model} & \textbf{Macro-F1} & \textbf{Precision} & \textbf{Recall} & \textbf{Macro-F1} & \textbf{Precision} & \textbf{Recall} \\
    \hline
        \multicolumn{7}{|c|}{\textbf{Transformers (fine-tuned)}} \\
    \hline
        longformer-large & 0.603 & 0.610 & 0.596 & 0.667 & \textbf{0.671} & 0.660 \\
        RoBERTa-large    & 0.595 & 0.601 & 0.585 & 0.664 & 0.669 & 0.652 \\
        BART-large       & 0.603 & 0.597 & 0.611 & \textbf{0.670} & 0.652 & \textbf{0.687} \\
        DeBERTa-large    & 0.601 & 0.598 & 0.606 & 0.661 & 0.657 & 0.669 \\
    \hline
        \multicolumn{7}{|c|}{\textbf{Proprietary LLMs (zero-shot)}}     \\
    \hline
        GPT-4            & 0.536 & 0.545 & 0.546 & 0.584 & 0.592 & 0.571 \\
        Claude-Opus      & 0.504 & 0.492 & 0.605 & 0.579 & 0.565 & 0.594 \\
    \hline
        \multicolumn{7}{|c|}{\textbf{Open-source LLMs (fine-tuned)}}    \\
    \hline
        LLaMa-3-8B       & 0.643 & 0.622 & 0.653 & -     & -     & -     \\
        Mistral-7B       & 0.637 & 0.621 & 0.646 & -     & -     & -     \\
        Gemma-7B         & 0.639 & 0.624 & 0.644 & -     & -     & -     \\  
    \hline    
    \end{tabular}
    \caption{performance of different approaches on Dev set : Task 3}
    \label{table:4}
\end{table*}

\begin{table*}[!ht]
    \centering
    \begin{tabular}{|l|ccc|ccc|}
    \hline
          & \multicolumn{3}{|c|}{\textbf{Direct Classification}} & \multicolumn{3}{|c|}{\textbf{And-rule Classification}} \\
    \hline
        \textbf{Model} & \textbf{Class1-F1} & \textbf{Precision} & \textbf{Recall} & \textbf{Class1-F1} & \textbf{Precision} & \textbf{Recall} \\
    \hline
          \multicolumn{7}{|c|}{\textbf{Transformers (fine-tuned)}} \\
    \hline
        longformer-large & 0.937 & \textbf{0.940} & 0.933 & -     & -     & -     \\
        RoBERTa-large    & 0.926 & 0.926 & 0.926 & -     & -     & -     \\
        BART-large       & \textbf{0.940} & 0.933 & 0.947 & -     & -     & -     \\
        DeBERTa-large    & 0.927 & 0.914 & 0.941 & -     & -     & -     \\
    \hline
         \multicolumn{7}{|c|}{\textbf{Proprietary LLMs (zero-shot)}}     \\
    \hline
        GPT-4            & 0.786 & 0.862 & 0.956 & 0.859 & 0.785 & 0.948 \\
        Claude-Opus      & 0.689 & 0.809 & \textbf{0.985} & 0.851 & 0.782 & 0.943 \\
    \hline
         \multicolumn{7}{|c|}{\textbf{Open-source LLMs (fine-tuned)}}    \\
    \hline
        LLaMa-3-8B       & 0.925 & 0.939 & 0.911 & -     & -     & -     \\
        Mistral-7B       & 0.921 & 0.921 & 0.921 & -     & -     & -     \\
        Gemma-7B         & 0.920 & 0.934 & 0.907 & -     & -     & -     \\  
    \hline    
    \end{tabular}
    \caption{performance of different approaches on Dev set : Task 5}
    \label{table:5}
\end{table*}

\begin{table*}[!ht]
    \centering
    \begin{tabular}{|l|ccc|ccc|}
    \hline
          & \multicolumn{3}{|c|}{\textbf{Direct Classification}} & \multicolumn{3}{|c|}{\textbf{Or-rule Classification}} \\
    \hline
        \textbf{Model} & \textbf{Class1-F1} & \textbf{Precision} & \textbf{Recall} & \textbf{Class1-F1} & \textbf{Precision} & \textbf{Recall} \\
    \hline
          \multicolumn{7}{|c|}{\textbf{Transformers (fine-tuned)}} \\
    \hline
        longformer-large & 0.898 & 0.884 & 0.914 & -     & -     & -     \\
        RoBERTa-large    & 0.891 & 0.862 & 0.920 & -     & -     & -     \\
        BART-large       & \textbf{0.901} & 0.878 & 0.926 & -     & -     & -     \\
        DeBERTa-large    & 0.894 & 0.869 & 0.923 & -     & -     & -     \\
    \hline
         \multicolumn{7}{|c|}{\textbf{Proprietary LLMs (zero-shot)}}     \\
    \hline
        GPT-4            & 0.861 & 0.791 & \textbf{0.960} & 0.897 & 0.870 & 0.925 \\
        Claude-Opus      & 0.858 & 0.794 & 0.952 & 0.893 & 0.873 & 0.937 \\
    \hline
         \multicolumn{7}{|c|}{\textbf{Open-source LLMs (fine-tuned)}}    \\
    \hline
        LLaMa-3-8B       & 0.898 & \textbf{0.912} & 0.886 & -     & -     & -     \\
        Mistral-7B       & 0.894 & 0.908 & 0.883 & -     & -     & -     \\
        Gemma-7B         & 0.894 & 0.901 & 0.889 & -     & -     & -     \\  
    \hline    
    \end{tabular}
    \caption{performance of different approaches on Dev set : Task 6}
    \label{table:6}
\end{table*}

\section{Error Analysis}
The LLMs performed equally good on all kinds of data while transformers models performed less effectively when the kind of language used is off from rest of the data or when criteria for classification was mentioned in one sentence and referred to the conditions indirectly later on. It was observed that positively labelled samples were predicted correctly by either the LLM approach or transformers, hence ensembles of both had recall over 0.99 with just 1 percent drop in F1 scores in Task 5 and 6. Many of the positively misclassified samples were in the format of advertisements where the title appears to match the criteria for positive classification. This is one area where LLMs were still able to distinguish effectively while other models did not.

\section{Conclusion}
the performance of some of the models compared to others on the test set can be seen in \autoref{table:7}, \autoref{table:8} and \autoref{table:9}. The LLM approach did yield comparatively good results despite using in a 4bit precision due to lack of computational resources. It is likely the performance would be better that the current models in full precision.  Many of the positive label texts have been filtered out during the data collection process. For example, texts self-reporting age in text format instead of numerical. Due to this, a higher focus on recall is necessary. A custom metric with higher importance to recall is better suited for Task 5 and 6 compared to F1 scores. Ensemble approaches like majority voting and filtering guaranteed positive label texts using LLM predictions could improve performance without a significant drop in the F1 scores. Finally, the performance improved on all the tasks while using dev set as additional training data compared to just the training data, hinting at the possibility of improving the performance by adding more training data. Augmentation through paraphrasing existing data however did not improve the results. 

\nocite{kingma2017adammethodstochasticoptimization,PMID:37986776}
\bibliography{anthology,custom}
\bibliographystyle{acl_natbib}
\clearpage

\appendix
\label{sec:appendix}

\section{Task 3 System Overview}
\begin{figure*}[t]
    \centering
    \includegraphics[width=1\linewidth]{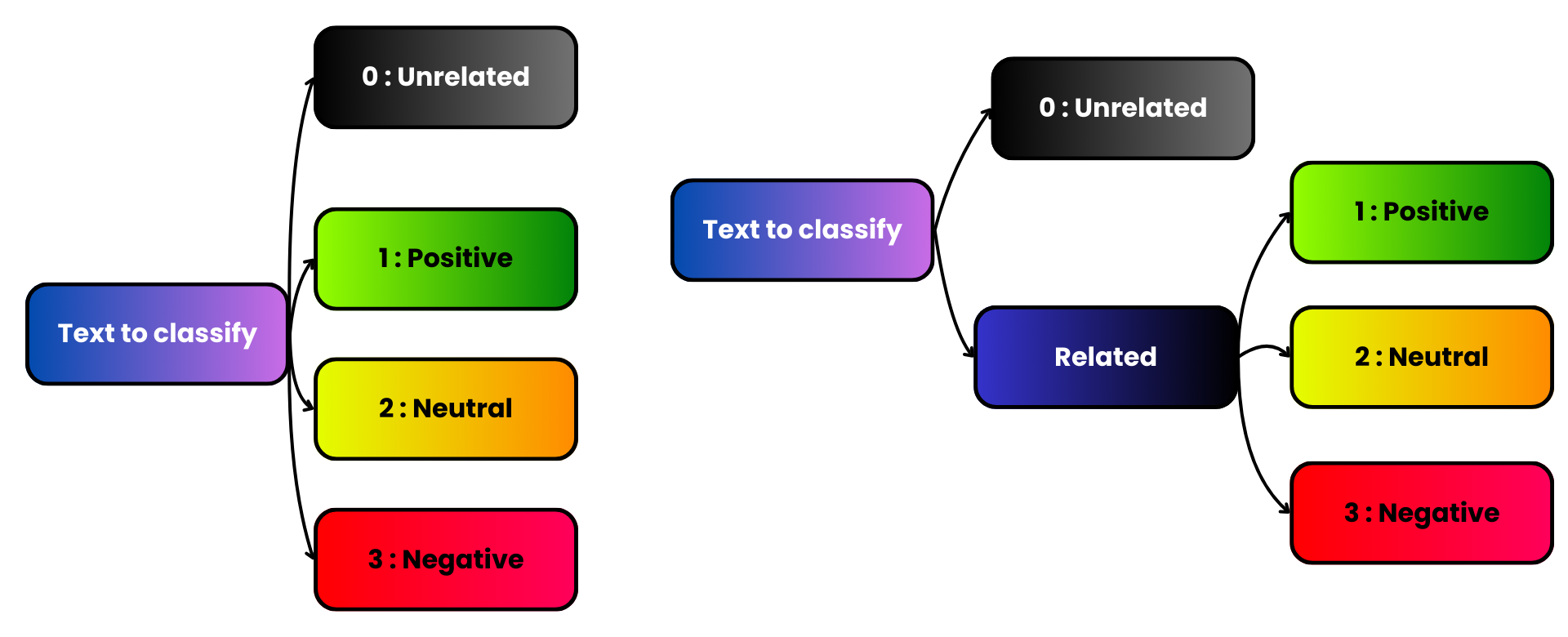}
    \caption{Overview of approaches used for Task 3 : Direct (left) and 2-Stage (right)}
    \label{fig:Task3}
\end{figure*}

Classifying the class of unrelated texts (class 0) from the other 3 separately had improved the performance by reducing mis-classification between Class 0 and others. The overview of the process can be seen in \autoref{fig:Task3}. The fine-tuned transformer models used had the best results with a learning rate of 0.00002 and weight decay of 0.01 over 30 epochs for 2-stage classification and 50 epochs for direct classification. In case of the fine-tuned LLMs, the base models were loaded in 4-bit configuration due to computational limitations, later fine-tuned and used in 16-bit precision for inference. During training, RoPE scaling was used for texts longer than 2048 tokens. They were fine-tuned over 3 epochs with a learning rate of 0.0002 and weight decay of 0.01 using Alpaca prompts.\\

The prompts ussed over the LLMs were as follows :
\begin{itemize}
    \item \textbf{2-stage 1st prompt} : "Did outdoor spaces or activities get mentioned? Respond only with a 1 for yes or 0 for no. Only one character (0/1) nothing else." 
    \item \textbf{2-stage 1st prompt} : "What impact did outdoor spaces or activities have on the user's mental health ? Respond only with a 1 for positive or 2 for neutral or 3 for negative. Only one character (1/2/3) nothing else."
    \item \textbf{Direct classification prompt} : "What impact did outdoor spaces or activities have on the user's mental health ? Respond only with a 1 for positive or 2 for neutral or 3 for negative or 0 for no mention. Only one character (1/2/3/0) nothing else."
    \item \textbf{Fine-tuned LLMs prompt} : "What impact did outdoor spaces or activities have on the user's mental health ? Respond only with a 1 for positive or 2 for neutral or 3 for negative or 0 for no mention. Only one character (1/2/3/0) nothing else"
\end{itemize}

The models that resulted in the best performance on the test set are available at :
\begin{itemize}
    \item \href{https://huggingface.co/1024m/SMM4H-Task3-BartL-1A30}{https://huggingface.co/1024m/SMM4H-Task3-BartL-1A30}
    \item \href{https://huggingface.co/1024m/SMM4H-Task3-BartL-1B30}{https://huggingface.co/1024m/SMM4H-Task3-BartL-1B30}
\end{itemize}

\section{Task 5 System Overview}
\begin{figure*}[t]
    \centering
    \includegraphics[width=1\linewidth]{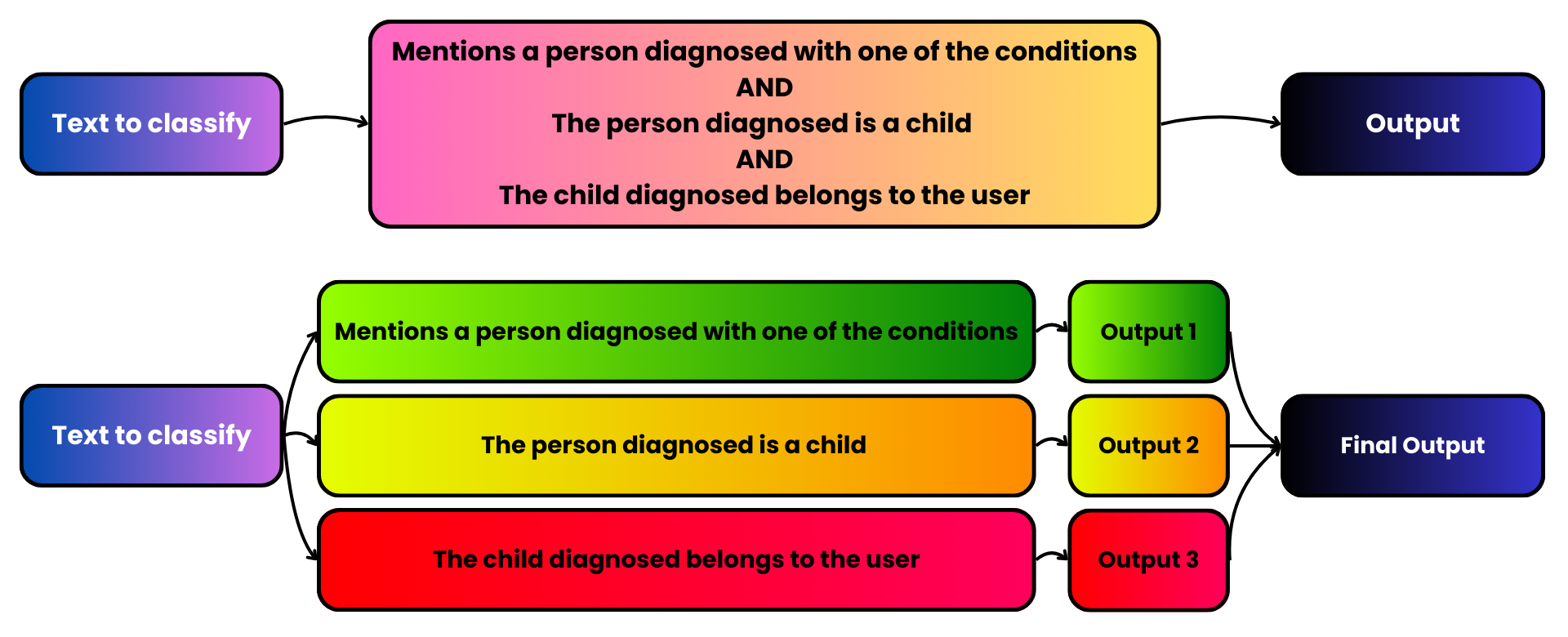}
    \caption{Overview of approaches used for Task 5 : Direct (top) and AND-rule (bottom)}
    \label{fig:Task5}
\end{figure*}

The overview of the process can be seen in \autoref{fig:Task5}. The fine-tuned transformer models used had the same hyper-parameters as used in Task 3, and were fine-tuned over 20 epochs. In case of the fine-tuned LLMs, the process is same as what was used in task 3. The proprietary systems were tested additionally using multiple separate prompts for each sub-condition that is to be true to be classified as a positive class text. In case of And-rule approach, the texts were marked as positive (class 1) if all of the conditions were met to achieve higher F1 with a lower recall trade-off.

The prompts used over the LLMs were as follows :
\begin{itemize}
    \item \textbf{Direct classification prompt} : "The tweets already mention at least one of the following: attention-deficit/hyperactivity disorder (ADHD), autism spectrum disorders (ASD), delayed speech (speech disorder), or asthma. In some cases, the tweets discuss hypothetical cases or the possibility of having the condition. It might be about someone else's child or an adult son/daughter. Respond with '1' if the tweet explicitly mentions an existing formal diagnosis of one of those conditions AND it concerns a child/baby AND the child is the user's own. In all other cases, respond with a '0'. Respond with only one character ('0'/'1') and nothing else."
    \item \textbf{AND-rule prompt 1} : "The tweets already mention at least one of the following: attention-deficit/hyperactivity disorder (ADHD), autism spectrum disorders (ASD), delayed speech (speech disorder), or asthma. In some cases, the tweets discuss hypothetical cases or the possibility of having the condition. Respond with '1' if the tweet explicitly mentions an existing formal diagnosis of one of those conditions. In all other cases, respond with a '0'. Respond with only one character ('0'/'1') and nothing else."
    \item \textbf{AND-rule prompt 2} : "The tweets already mention... ...Respond with '1' if the tweet explicitly mentions it concerns a child/baby having one of those conditions. In all other cases, respond with a '0'. Respond with only one character ('0'/'1') and nothing else."
    \item \textbf{AND-rule prompt 3} : "The tweets already mention... ...Respond with '1' if the tweet explicitly mentions the child is the user's own having diagnosed with one of those conditions. In all other cases, respond with a '0'. Respond with only one character ('0'/'1') and nothing else."
\end{itemize}

The model that resulted in the best performance on the test set is available at :
\begin{itemize}
    \item \href{https://huggingface.co/1024m/SMM4H-Task5-BartL-2A}{https://huggingface.co/1024m/SMM4H-Task5-BartL-2A}
\end{itemize}

\section{Task 6 System Overview}
\begin{figure*}[t]
    \centering
    \includegraphics[width=1\linewidth]{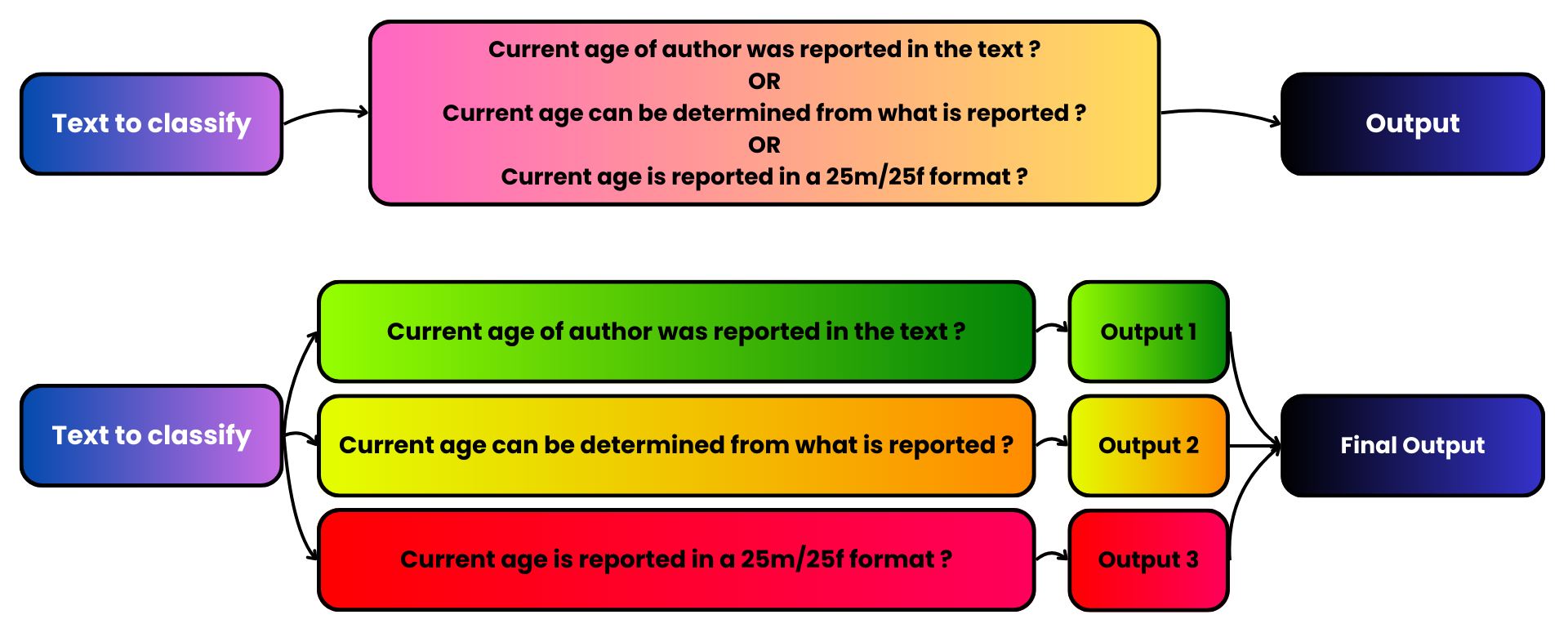}
    \caption{Overview of approaches used for Task 6 : Direct (top) and OR-rule (bottom)}
    \label{fig:Task6}
\end{figure*}

The overview of the process can be seen in \autoref{fig:Task6}. The fine-tuned transformer models used had the same hyper-parameters as used in Task 3, and were fine-tuned over 20 epochs. In case of the fine-tuned LLMs, the process is same as what was used in task 3. The proprietary systems were tested additionally using multiple separate prompts for each sub-condition that can be true to be classified as a positive class text. In case of OR-rule approach, the texts were marked as positive (class 1) if at least one of the conditions were met to achieve higher F1 with a lower recall trade-off. The classification was done separately fro twitter and reddit posts with separate models i.e one for each platform's posts. 

The prompts used over the LLMs were as follows :
\begin{itemize}
    \item \textbf{Direct classification prompt} : "Respond only with 0 or 1 and nothing else : based on whether current age of the AUTHOR in years can be known from the texts. The texts have a two digit number which is likely an age if not clear. The age needed to know in context is current age of THE author and not someone else. In some cases formats like 25m , 24f are used where m refers to Male and f refers to Female."
    \item \textbf{OR-rule prompt 1} : "Respond only with 0 or 1 and nothing else based on whether the current age of the author was reported in the given text."
    \item \textbf{OR-rule prompt 2} : "Respond only with 0 or 1 and nothing else based on whether the current age of the author can be determined from the given text."
    \item \textbf{OR-rule prompt 3} : "Respond only with 0 or 1 and nothing else based on whether the current age of the author was expressed using formats like 25m , 24f are used where 'm' refers to Male and 'f' refers to Female."
\end{itemize}

The models that resulted in the best performance on the test set are available at :
\begin{itemize}
    \item \href{https://huggingface.co/1024m/SMM4H-Task6-BartL-A20}{https://huggingface.co/1024m/SMM4H-Task6-BartL-A20} For Reddit texts
    \item \href{https://huggingface.co/1024m/SMM4H-Task6-BartL-B20}{https://huggingface.co/1024m/SMM4H-Task6-BartL-B20} For Twitter texts
\end{itemize}






\end{document}